\pdfoutput=1

\documentclass[11pt]{article}

\usepackage[]{ACL2023}

\usepackage{times}
\usepackage{latexsym}

\usepackage[T1]{fontenc}

\usepackage[utf8]{inputenc}

\usepackage{microtype}

\usepackage{inconsolata}

\usepackage{bbm}
\usepackage{pifont}
\usepackage{array}

\usepackage{subfig}
\usepackage{multirow}
\usepackage{multicol}
\usepackage{arydshln}
\usepackage{setspace}
\usepackage{booktabs}

\usepackage{amsmath, amsfonts}
\usepackage{graphicx}

\usepackage{algorithmicx}
\usepackage{algorithm}
\usepackage{algpseudocode}

\makeatletter
\newcommand{\removelatexerror}{\let\@latex@error\@gobble}
\makeatother

\newcommand{\tabincell}[2]{
\begin{tabular}{@{}#1@{}}#2\end{tabular}
}

\algnewcommand{\LeftComment}[1]{\State \(\triangleright\) #1}

\usepackage{xcolor}

\title{When Gradient Descent Meets Derivative-Free Optimization: A Match Made in Black-Box Scenario}

\author{
Chengcheng Han\textsuperscript{$\diamondsuit$}\thanks{~~Equal contribution.}\quad \quad 
Liqing Cui\textsuperscript{$\diamondsuit$}\footnotemark[1]\quad \quad 
Renyu Zhu\textsuperscript{$\diamondsuit\spadesuit$}\\
\bf{
Jianing Wang\textsuperscript{$\diamondsuit$}\quad
Nuo Chen\textsuperscript{$\diamondsuit$}\quad
Qiushi Sun\textsuperscript{$\diamondsuit\heartsuit$}\quad
Xiang Li\textsuperscript{$\diamondsuit$} \quad
Ming Gao\textsuperscript{$\diamondsuit\clubsuit$}\thanks{~~Corresponding author.}
}\\
\textsuperscript{$\diamondsuit$}School of Data Science and Engineering\thanks{~~Shanghai Engineering Research Center of Big Data Management}, East China Normal University\\
\textsuperscript{$\spadesuit$}NetEase Fuxi AI Lab\\
\textsuperscript{$\heartsuit$}Department of Mathematics, National University of Singapore\\
\textsuperscript{$\clubsuit$}KLATASDS-MOE, School of Statistics, East China Normal University\\
\texttt{\{chengchenghan,liqingcui,jianingwang,nuochen,qiushisun\}@stu.ecnu.edu.cn}\\
\texttt{zhurenyu@corp.netease.com}\\
\texttt{\{xiangli,mgao\}@dase.ecnu.edu.cn}
}

\begin{document}
\maketitle
\begin{abstract}
Large pre-trained language models~(PLMs)
have garnered significant attention 
for their versatility 
and potential 
for solving a wide spectrum of natural language processing~(NLP) tasks.
However,
the cost of running these PLMs may be prohibitive.
Furthermore,
PLMs may not be open-sourced
due to commercial considerations
and potential risks of misuse,
such as GPT-3.
The parameters 
and gradients of PLMs 
are unavailable in this scenario.
To solve the issue,
black-box tuning has been proposed,
which utilizes derivative-free optimization~(DFO),
instead of gradient descent,
for training task-specific continuous prompts.
However,
these gradient-free methods
still exhibit a significant gap
compared to gradient-based methods.
In this paper,
we introduce gradient descent into 
black-box tuning scenario 
through knowledge distillation.
Furthermore,
we propose a novel method GDFO,
which integrates gradient descent
and derivative-free optimization 
to optimize task-specific continuous prompts
in a harmonized manner.
Experimental results show that
GDFO can achieve significant performance gains
over previous state-of-the-art methods.

\end{abstract}

\section{Introduction}
\label{Intro}

Large pre-trained language models~(PLMs)~\cite{devlin2019bert, liu2019roberta, yang2019xlnet, raffel2020_t5} have attracted considerable attention 
for their versatility 
and potential 
for solving a wide spectrum of Natural Language Processing~(NLP) tasks.
Especially,
through prompt-based learning~(PL)~\cite{liu2021ptuningv2, gu2022ppt},
PLMs have consistently demonstrated impressive performance 
on various downstream tasks with a few labeled samples.
However,
it is a challenge to extend the benefits of these large PLMs
to a broader audience.
For users,
the cost of running these models may be prohibitive;
for service providers,
they may not open source the model parameters
due to commercial considerations
and potential risks of misuse\footnote{\url{https://openai.com/blog/openai-api/}}.
One possible solution is to deploy PLMs as a service,
enabling users to access the advanced capabilities of PLMs 
through their inference APIs, 
such as 
GPT-3~\cite{brown2020GPT3}, 
ERNIE~\cite{sun2021ernie3}
and Yuan~\cite{wu2021yuan1}.

\begin{figure}[!t]
    \centering
    \includegraphics[width=0.48\textwidth]{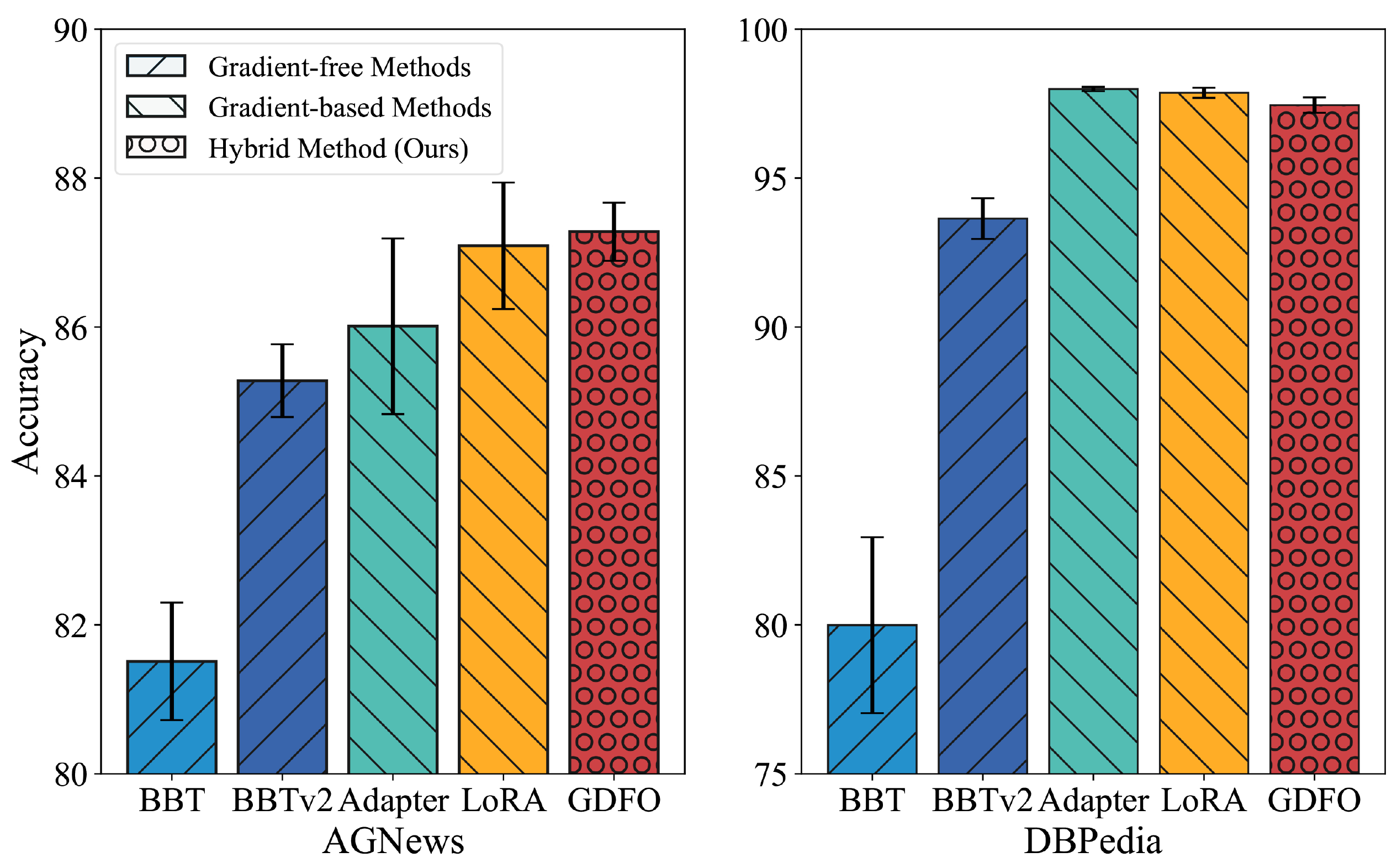}
    \caption{Accuracy~(\%) on the AG's News and DBPedia datasets.
    Experimental setup is detailed in Section~\ref{sec:exp_setting}.
    It is noted that 
    prior gradient-based approaches, 
    such as Adapter~\cite{houlsby2019adapter} 
    and LoRA~\cite{hu2021lora},
    are unable to be used in black-box scenarios.
    GDFO is the first to introduce gradient descent 
    in the black-box tuning scenario.}
    \label{fig:intro}
\end{figure}

In this scenario,
the large pre-trained language model 
provided by the server is considered as a black box. 
In order to
perform various downstream tasks,
users are required to construct task-specific prompts 
or select training samples~\cite{brown2020GPT3} 
to input into the black box.
We can manually construct discrete prompts,
which are simple and effective
but may not fully utilize training data,
potentially resulting in suboptimal performance
on some tasks.
Instead of designing hand-crafted discrete prompts,
there have been 
an increasing number of studies 
on continuous prompt tuning~\cite{lester2021ptuning,liu2021ptuningv2,ding2022prompttuning2},
which aim to train continuous prompts and 
add them to the original samples.
Trainable continuous prompts have also shown remarkable success
on various tasks,
but most existing methods 
optimize the continuous prompts through back-propagation,
which is unavailable in the black-box scenario.
To solve the issue,
\citet{sun2022bbt} have recently
proposed Black Box Tuning~(BBT),
which utilizes random projection matrices 
and derivative-free optimization~(DFO)~\cite{kolda2003optimizationDFO1,conn2009introductionDFO2,rios2013derivativeDFO3},
instead of gradient descent,
for training continuous prompts in the black-box scenario.
Built upon BBT,
BBTv2~\cite{sun2022bbtv2} prepends continuous prompts to each layer of the PLM 
and further presents a 
divide-and-conquer gradient-free algorithm 
to alternately optimize the prompts 
at different layers.
Both 
BBT and BBTv2 
have shown their superiority against 
other gradient-free methods. 
Despite the success, 
there remains a significant gap 
compared to gradient-based methods on certain tasks.
For example,
compared against BBTv2,
Adapter~\cite{houlsby2019adapter},
a gradient-based method,
leads by $4.35\%$
on the DBPedia dataset~(as shown in Figure~\ref{fig:intro}).
Therefore,
we consider that 
the incorporation of gradient descent 
into the black-box scenario 
may potentially enhance
the performance of the model.

Based on the insights discussed above,
in this paper,
we introduce gradient descent into the black-box scenario 
through knowledge distillation techniques.
In particular, 
we propose a novel approach
named \textbf{GDFO}
to combine 
\textbf{G}radient descent with
\textbf{D}erivative-\textbf{F}ree \textbf{O}ptimization, 
allowing them to 
jointly optimize task-specific continuous prompts.
First,
we adopt the technique of knowledge distillation,
where
a \emph{student model} is trained 
to emulate the knowledge of the black-box model,
referred to as the \emph{teacher model}. 
Then,
a prompt generator is trained by gradient descent
through the \emph{student model}, 
while utilizing derivative-free optimization algorithms 
to optimize continuous task-specific prompts.
The continuous prompts generated by the prompt generator
and the prompts optimized by the derivative-free algorithm 
are further integrated to serve as the final prompts.
Finally,
we perform extensive experiments on seven benchmark datasets to
show that 
GDFO
can achieve significant performance gains
over other state-of-the-art methods.
The main contributions of the paper are summarized as follows:

\begin{itemize}

\item 
To the best of our knowledge, 
we are the first to utilize gradient descent 
to optimize task-specific continuous prompts 
in the black-box scenario
through knowledge distillation.

\item
We propose a novel method~GDFO,
which integrates 
gradient descent and derivative-free optimization
to optimize task-specific continuous prompts 
in a harmonized manner.

\item
We conduct comprehensive experiments
on seven benchmark datasets 
under the black-box scenario.
Empirical results demonstrate
the superiority of GDFO over other competitors.
\end{itemize}

\begin{figure*}[htbp]
    \centering
    \includegraphics[height=250pt, width=455pt]{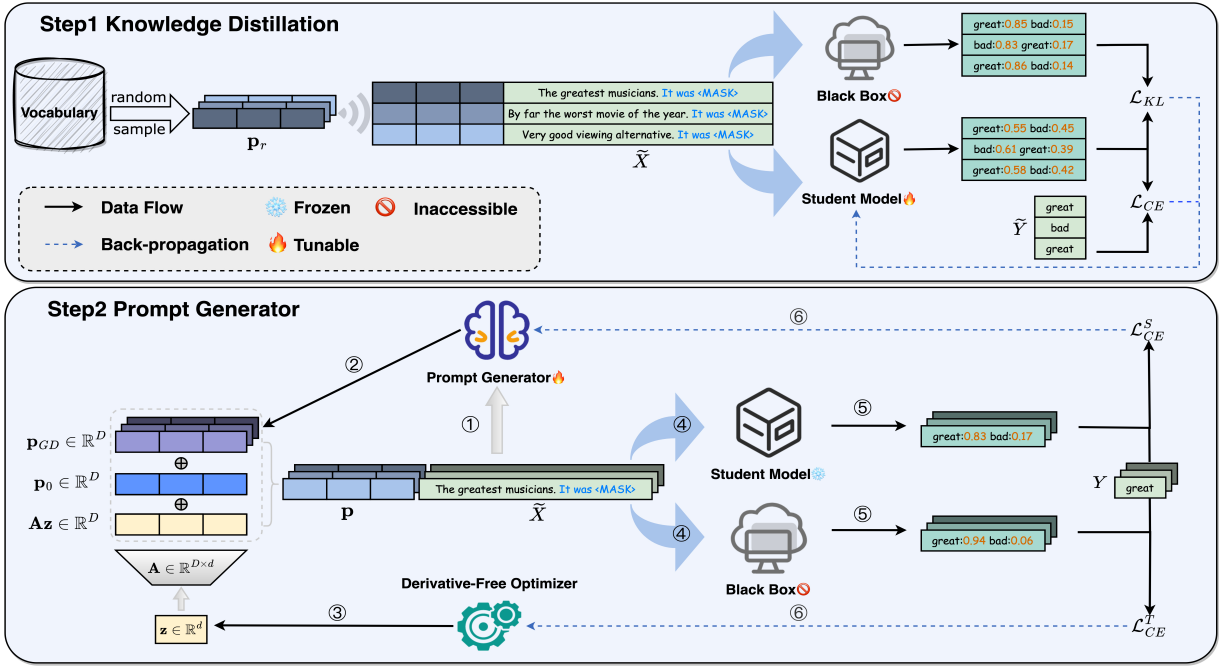}
    \caption{The overall architecture of GDFO.
    The details of the model are described in Section~\ref{sec:method}. The training procedure is shown in Algorithm~\ref{alg:train}.}
    \label{fig:framework}
\end{figure*}

\section{Related Work}
\subsection{Prompt-based Learning}
Prompt-based learning, 
in which the PLM is adapted to various tasks 
by task-specific prompts, 
has emerged as a promising framework.
\citet{brown2020GPT3} shows that
PLM can perform excellently in few-shot learning
by using manual prompts concatenated with samples.
However, designing prompts in a hand-crafted fashion 
requires substantial time and experience 
and may not find the optimal ones~\cite{jiang2020howcanwe, shin2021constrained}.
To solve the problem, 
researchers attempt to use automated prompts.
LM-BFF~\cite{gao2021lmbff}
uses prompt-based fine-tuning 
with automatically searched prompts
and generates task demonstrations 
to be a part of the input context.
P-tuning~\cite{liu2021ptuning} 
optimizes the continuous prompts using gradient descent 
as an alternative to discrete prompt searching.
P-tuning v2~\cite{liu2021ptuningv2} adopts continuous prompts for each layer of the PLMs
to improve the model performance.
Prefix-tuning~\cite{li2021prefixtuning} optimizes continuous task-specific vectors
and prepends them to the input texts. 
Input-tuning~\cite{an2022inputtuning} fine-tunes both
the continuous prompts and
the input representations,
leading to a more effective way to
adapt unfamiliar inputs to frozen PLMs.
\subsection{Black-box Tuning}
Due to commercial considerations,
large PLMs such as GPT-3~\cite{brown2020GPT3}
are only provided as a service
in the cloud, 
resulting inaccessible parameters and gradients of PLMs.
To tackle this issue,
BBT~\cite{sun2022bbt,diao2022black}
has been proposed 
to optimize the continuous prompts
via derivative-free optimization~(DFO). 
As an improved version of BBT,
BBTv2~\cite{sun2022bbtv2} inserts prompts to each layer of the PLMs
instead of optimizing the prompt merely in the input layer.
Furthermore, 
GrIPS~\cite{prasad2022grips} proposes a gradient-free search approach
to generate discrete prompts.
Besides, 
RLPrompt~\cite{deng2022rlprompt} optimizes discrete prompts 
through reinforcement learning
and utilizes a continuous policy network 
which is highly parameter-efficient to generate prompts. 
PALP~\cite{cho2022prompt} combines linear models 
and in-context learning~\cite{brown2020GPT3}
to augment training samples with the templates 
for better contextualization. 
To improve the computational efficiency,
PromptBoosting~\cite{hou2022promptboosting} 
constructs a pool of prompts
via a gradient-free approach
and 
ensembles many weak learners
using the ADABOOST algorithm
to enhance the model performance.
Despite the success of the above approaches,
all of them do not optimize continuous prompts
through gradient descent~(GD) in the black-box scenario,
our method introduces GD to the scenario
through knowledge distillation 
and combines GD and DFO to jointly
optimize continuous prompts,
which provides a novel insight
for future black-box tuning approaches.

\subsection{Knowledge Distillation}
As a representative method of model compression,
knowledge distillation 
transfers the knowledge from a larger deep neural network~(\emph{teacher}) 
to a smaller network~(\emph{student})~\cite{hinton2015kd, kim2016sequence}. 
There have been different distillation algorithms being proposed 
to face more complex settings of transferring knowledge, 
including adversarial distillation~\cite{ma2020adversarial, wang2022blackboxkd}, 
multi-teacher distillation~\cite{guan2020multiteacherkd, yuan2021multiteacherkd}
and
data-free distillation~\cite{fang2022datafreekd, binici2022robustdatafreekd}. 
Furthermore, the superior success of PLMs has also spurred researchers 
to distill PLMs into smaller models 
while retaining performance.
DistilBERT~\cite{sanh2019distilbert} introduces
a triple loss combining language modeling
and cosine-distance losses 
to leverage the inductive biases
learned by large models during pre-training. 
TinyBERT~\cite{jiao2020tinybert} performs a Transformer distillation method 
at both the pre-training and task-specific learning stages. 
NewsBERT~\cite{wu2021newsbert} designs a collaborative learning framework 
where the \emph{student model} can learn 
from the experience of the \emph{teacher model}. 
In this paper, we consider knowledge distillation
to transfer knowledge 
from a black-box \emph{teacher} to a \emph{student},
which is used for training a prompt generator by gradient descent.

\section{Method}
\label{sec:method}

In this section, 
we describe our approach~GDFO.
We first give an overview of GDFO,
which is illustrated in Figure~\ref{fig:framework}.
GDFO first trains a \emph{student model}
by aligning its outputs
to that of the \emph{teacher model}~(i.e., the black-box model).
Then,
GDFO trains the prompt generator by gradient descent
while simultaneously optimizing the continuous prompts via DFO.
Finally, 
the final prompts are obtained by 
integrating the prompts 
generated by the prompt generator 
with those optimized by DFO,
which are then
fed into the black-box model 
together with query instances
to obtain predictions.
Next, we describe 
each component of GDFO in detail.

\subsection{Knowledge Distillation}

Given a student model $S$ 
and a teacher model $T$,
the objective of knowledge distillation~(KD)
is to enhance the performance of $S$
by aligning its outputs with 
those of $T$,
which is accomplished 
by reducing the divergence between 
the probability distributions
generated by $S$ and $T$.
In the black-box scenario,
the black-box model is considered as $T$.
We utilize $T$'s outputs as 
soft targets for $S$ to learn.
Given a training instance,
we randomly select $n$ tokens
from the PLM vocabulary 
to construct a random prompt~$\mathbf{p}_r$,
which is concatenated to
the beginning of the instance.
Additionally,
a hand-crafted template\footnote{The details of templates are shown in Table~\ref{tab:datasets}.}
is appended to the end of the instance.
We use the concatenated sentence 
as the input $x$.
We denote $S(x)$ and $T(x)$
as the output logits of
$S$ and $T$ for input $x$,
respectively.
The KD can be conducted 
by minimizing the Kullback-Leibler~(KL)
divergence distance between
the \emph{student} and \emph{teacher} predictions:
\begin{eqnarray}
\label{eq:KL}
\mathcal{L}_{KL}=\operatorname{KL}(\sigma(S(x) / \tau) \| \sigma(T(x) / \tau)),
\end{eqnarray}
where $\sigma(\cdot)$ denotes the \texttt{softmax} function and $\tau$ is a temperature hyper-parameter.
The \emph{student} parameters are updated 
according to $\mathcal{L}_{KL}$ 
and the cross-entropy loss~$\mathcal{L}_{CE}$
over the ground-truth $y$:
\begin{eqnarray}
\label{eq:KL+CE}
\mathcal{L}=\left(1-\lambda \right) \mathcal{L}_{CE}+\lambda \mathcal{L}_{KL},
\end{eqnarray}
where $\lambda$ is a weight 
and $\mathcal{L}_{CE}$ is defined as:
\begin{equation}
\label{eq:CE}
\mathcal{L}_{CE}=-y \log \sigma(S(x)).
\end{equation}

\subsection{Prompt Generator}

\begin{figure}[!t]
  \begin{algorithm}[H]
    \caption{Training Procedure}
    \label{alg:train}
    \begin{algorithmic}[1]
      \Require Training data~$ \{\mathcal{X}_{train}, \mathcal{Y}_{train}\} $;
              Black-box model~$T$;
              Student model~$S_{\theta}$;
              Prompt generator~$G_{\mu}$;
              Epochs for knowledge distillation~$E_{kd}$;
              The number of API calls~$N$;
              The PLM vocabulary~$\mathcal{V}$;
              Hand-crafted template~$t$;
      
      \Statex \# Knowledge Distillation
    
    
      \For{each $i \in E_{kd}$}
        \For{each $x \in \mathcal{X}_{train}$}
          \State $\mathbf{p}_r \leftarrow \texttt{Random\_Sample}(\mathcal{V}, n)$;
          \State $\hat{y}_T, \hat{y}_S \leftarrow T([\mathbf{p}_r;x;t]), S_{\theta}([\mathbf{p}_r;x;t])$;
          \State Calculate $\mathcal{L}$ by Equation~\ref{eq:KL+CE};
          \State Update $\theta$ by $\mathcal{L}$;
        \EndFor
      \EndFor
      \Statex \# Prompt Generator
      
      \For{each $i \in N$}
        \For{each $x \in \mathcal{X}_{train}$}
            \State $\mathbf{p}_{GD} \leftarrow G_{\mu}(x)$;
            \State Get $\mathbf{z}$ by CMA-ES;
            \State Get $\mathbf{p}$ by Equation~\ref{eq:P};
            \State $\hat{y}_T, \hat{y}_S \leftarrow T([\mathbf{p};x;t]), S([\mathbf{p};x;t])$;
            \State Calculate $\mathcal{L}^{T}_{CE}, \mathcal{L}^{S}_{CE}$ by Equation~\ref{eq:CE};
            \State Update $\mu$ by $\mathcal{L}^{S}_{CE}$;
            \State Optimize CMA-ES by $\mathcal{L}^{T}_{CE}$;
        \EndFor
      \EndFor
    \end{algorithmic}
  \end{algorithm}
\end{figure}

Upon the completion 
of training $S$
via knowledge distillation,
the \emph{student} parameters are frozen 
and a prompt generator is optimized
by gradient descent
with the purpose of generating 
continuous prompts~$\mathbf{p}_{\text{\tiny GD}}\in \mathbb{R}^D$ 
for given samples.
Meanwhile,
following BBT~\cite{sun2022bbt},
we optimize intermediate vector~$\mathbf{z}\in \mathbb{R}^d$
through CMA-ES~(Covariance Matrix Adaptation Evolution Strategy)~\cite{hansen2001cmaes1,hansen2003cmaes2}, 
which is a widely used evolutionary algorithm
for non-convex black-box optimization
in continuous domain.
Then a random projection matrix $\mathbf{A}\in \mathbb{R}^{D\times d}$ 
is utilized to 
project $\mathbf{z}$ into the high-dimensional space.
Finally,
we randomly sample 
$n$ tokens from the PLM vocabulary as 
initial prompt $\mathbf{p}_0$
and get final continuous prompt~$\mathbf{p} \in \mathbb{R}^D$:
\begin{equation}
\label{eq:P}
\mathbf{p} = \alpha \mathbf{p}_{\text{\tiny GD}} + (1-\alpha)(\mathbf{p}_0 + \mathbf{Az}),
\end{equation}
where $\alpha$ is a balancing weight.
Further information 
regarding the initialization of $\mathbf{A}$ 
and the specifics 
of the optimization procedure 
of CMA-ES
can be found in~\citet{sun2022bbt}.
Given a training instance,
$\mathbf{p}$ is concatenated to
the beginning of it
and a hand-crafted template\footnotemark[2]
is appended to the end of it.
The concatenated sample
is fed into $S$ and $T$.
Then the output logits
are obtained 
and used to compute $\mathcal{L}_{CE}$,
which is utilized to update the parameters
of the prompt generator 
and optimize $\mathbf{z}$ through CMA-ES.
The overall training procedure of GDFO
is summarized in Algorithm~\ref{alg:train}.

\subsection{Inference}

\begin{figure}[htbp]
    \centering
    \includegraphics[width=0.48\textwidth]{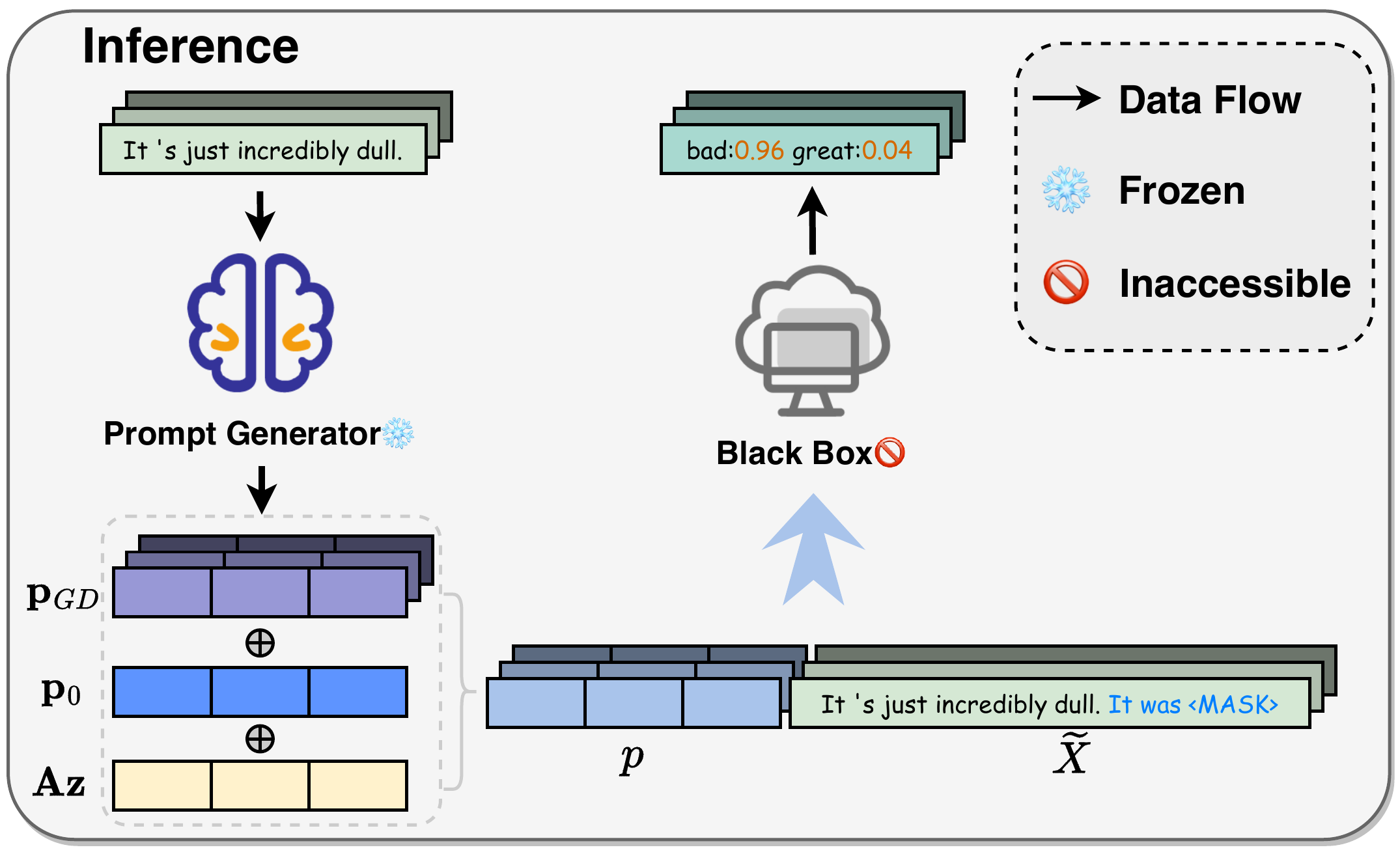}
    \caption{The inference procedure of GDFO.}
    \label{fig:inference}
\end{figure}

During the inference stage, 
given a query instance,
we first input it 
into the prompt generator 
to generate $\mathbf{p}_{\text{\tiny GD}}$.
Subsequently,
we combine $\mathbf{p}_{\text{\tiny GD}}$,
$\mathbf{p}_0$, and $\mathbf{Az}$ 
that have been optimized 
through CMA-ES 
to obtain the final continuous prompt~$\mathbf{p}$
through the Equation~\ref{eq:P}.
Next, 
similar to the training stage,
we concatenate $\mathbf{p}$
to the front of the query instance
and append the hand-crafted template\footnotemark[2]
to the end of it.
Finally,
we input the concatenated sample 
to the black-box model to obtain the prediction.
The overall inference procedure is shown in Figure~\ref{fig:inference}.

\section{Experiments}

In this section, 
we perform comprehensive experiments 
to compare our proposed model with
twelve competitive baselines 
on seven downstream tasks.

\subsection{Datasets}
We perform experiments 
on a variety of language understanding tasks,
including sentiment analysis,
topic classification,
natural language inference~(NLI),
and paraphrasing.
Statistics of these datasets are given in Table~\ref{tab:datasets}.
Specifically, we utilize the following datasets:

\paragraph{Sentiment analysis:} 
SST-2~\cite{socher2013SST2}
and Yelp polarity~\cite{zhang2015YelpAGNewsDBPedia}
consist of text samples 
with assigned sentiment labels (e.g. positive or negative).

\paragraph{Topic classification:} 
AG's News~\cite{zhang2015YelpAGNewsDBPedia} 
and DBPedia~\cite{zhang2015YelpAGNewsDBPedia}
contain text samples with pre-defined topics.

\paragraph{NLI: }
SNLI~\cite{Bowman_2015SNLI}
and RTE~\cite{wang2018glueRTE}
are composed of sentence pairs 
and the objective is to 
determine the relationship between the two sentences, 
such as entailment, contradiction and neutral.

\paragraph{Paraphrase: }
MRPC~\cite{dolan2005automaticallyMRPC}
contains sentence pairs 
and the goal is 
to recognize semantic equivalence between the two sentences.

\begin{table*}[t]
	\begin{center}
	\resizebox{2.08\columnwidth}{!}{
		\begin{tabular}{ccccccc}
			\hline
			\hline
			\specialrule{0em}{1pt}{1pt}
			Category
			&Datasets
			&\# Classes
			&\tabincell{c}{\# Training \\ samples}
			&\tabincell{c}{\# Test \\ samples}
			&Templates
			&Label words \\
            \hline
            \specialrule{0em}{1pt}{1pt}
			\multirow{6}{*}{\tabincell{c}{Single\\sentence}}
            & SST-2
            & 2
            & 32
            & 0.9k
            & \texttt{<Sentence>}. It was \texttt{[MASK]}.
            & great, bad
            \\
            \specialrule{0em}{1pt}{1pt}
            & Yelp P.
            & 2
            & 32
            & 38k
            & \texttt{<Sentence>}. It was \texttt{[MASK]}.
            & great, bad
            \\
            \specialrule{0em}{1pt}{1pt}
            & AG's News
            & 4
            & 64
            & 7.6k
            & \texttt{[MASK]} News: \texttt{<Sentence>}
            & World, Sports, Business, Tech
            \\
            \specialrule{0em}{1pt}{1pt}
            & DBPedia
            & 14
            & 224
            & 70k
            & [Category: \texttt{[MASK]}] \texttt{<Sentence>}
            & \tabincell{l}{Company, Education, Artist, Athlete, Office,\\Transportation, Building, Natural, Village,\\Animal, Plant, Album, Film, Written}
            \\
			\hline
			\specialrule{0em}{1pt}{1pt}
			\multirow{3}{*}{\tabincell{c}{Sentence\\pair}}
            & MRPC
            & 2
            & 32
            & 0.4k
            & \texttt{<Sentence$_1$>} ? \texttt{[MASK]} , \texttt{<Sentence$_2$>}
            & Yes, No
            \\
            \specialrule{0em}{1pt}{1pt}
            & RTE
            & 2
            & 32
            & 0.3k
            & \texttt{<Sentence$_1$>} ? \texttt{[MASK]} , \texttt{<Sentence$_2$>}
            & Yes, No
            \\
            \specialrule{0em}{1pt}{1pt}
            & SNLI
            & 3
            & 48
            & 9.8k
            & \texttt{<Sentence$_1$>} ? \texttt{[MASK]} , \texttt{<Sentence$_2$>}
            & Yes, Maybe, No
            \\
			\hline
			\hline
		\end{tabular}
        }
	\end{center}
	\caption{Statistics, hand-crafted templates and label words of datasets.}
	\label{tab:datasets}
	
\end{table*}

\subsection{Baselines}
We compare 
GDFO with twelve competitive methods,
which can be grouped into two categories:
gradient-based methods and gradient-free methods.

For gradient-based methods, we
consider six baselines:
\textbf{(1) Model Tuning} fine-tunes the entire PLM through training data.
\textbf{(2) Adapter}~\cite{houlsby2019adapter} is a new module 
added between layers of a PLM. 
The parameters of the original network 
remain fixed, 
yielding a high degree of parameter sharing.
\textbf{(3) BitFit}~\cite{zaken2022bitfit} is a sparse-finetuning method
where most of the network parameters are frozen
and only the bias-terms of the model (or a subset of them)
are being modified.
\textbf{(4) LoRA}~\cite{hu2021lora},
an efficient adaptation strategy,
allows us to train some dense layers in a neural network indirectly
by optimizing
rank decomposition matrices
of the dense layers’ change, 
while keeping the pre-trained weights frozen.
\textbf{(5) Prompt Tuning}~\cite{lester2021ptuning} 
freezes the entire PLM 
and only allows additional tunable tokens 
to be prepended to the input text.
\textbf{(6) P-Tuning v2}~\cite{liu2021ptuningv2}
applys continuous prompts 
for every layer of the PLM
instead of the mere input layer.

For gradient-free methods,
we also consider six baselines:
\textbf{(1) Manual Prompt}
conducts subsequent experiments using hand-crafted prompts
following the pre-defined templates in Table~\ref{tab:datasets}.
\textbf{(2) In-Context Learning}~\cite{brown2020GPT3} 
provides a few training examples for the model
to improve its capability of few-shot learning.
\textbf{(3) Feature-MLP} trains a two-layered MLP classifier
provided with embeddings encoded by the PLM.
\textbf{(4) Feature-BiLSTM} trains a bidirectional LSTM on 
the word representations
and connects it to a linear classifier.
\textbf{(5) BBT}~\cite{sun2022bbt}
optimizes the continuous prompt prepended to the input text via
derivative-free optimization~(DFO).
\textbf{(6) BBTv2}~\cite{sun2022bbtv2}
proposes a divide-and-conquer algorithm
to alternately optimize the prompt at each layer 
of the PLM.
Compared with BBT,
BBTv2 inserts prompts to each layer of the PLM
instead of optimizing the prompt merely 
in the input layer.

\subsection{Implementation}
\label{sec:exp_setting}

\paragraph{Few-shot setting}
We adopt the same procedure 
as described in previous studies~\cite{zhang2020revisiting, sun2022bbtv2}
to establish a true few-shot learning environment.
Specifically,
we randomly select $k$ samples per class
to create a $k$-shot training set~$\mathcal{D}_{train}$,
and form a development set~$\mathcal{D}_{dev}$ 
by randomly selecting another $k$ samples 
from the original training set,
resulting in $|\mathcal{D}_{train}| = |\mathcal{D}_{dev}|$.
We use the original development sets 
as our test sets $\mathcal{D}_{test}$.
For datasets 
that do not have development sets,
we use the original test sets. 
It is noted that $|\mathcal{D}_{test}| \gg |\mathcal{D}_{train}| = |\mathcal{D}_{dev}|$.

\paragraph{Experimental settings}
To compare with BBTv2~\cite{sun2022bbtv2},
we mainly 
use $\text{RoBERTa}_\text{{\footnotesize LARGE}}$~\cite{liu2019roberta}
as the black-box model.
For hyper-parameters,
we use the grid search 
to find the best for our model.
For knowledge distillation,
we use BERT{\footnotesize LARGE}~\cite{devlin2019bert}
as our \emph{student model}.
We set 
the temperature~$\tau$ to 1
and the balancing weight~$\lambda$ to $0.5$.
We fine-tune the \emph{student model}
for 2,000 epochs with the learning rate $1e-4$.
For prompt generator,
we use a fully connected layer
and set the dimensionality of the fully connected layer
to 1024.
The learning rate of the prompt generator is $1e-5$.
For CMA-ES,
following \citet{sun2022bbt},
we set the prompt length~$n$ to 50.
The dimensionality of $\mathbf{z}$ is set to 500
and the population size of CMA-ES is set to 20.
The balancing weight~$\alpha$ is set to $0.5$.
We train our prompt generator and run CMA-ES for 8,000 API calls.
All baseline results are recorded in \citet{sun2022bbtv2}.
We run all the experiments on a single NVIDIA v100 GPU.

\subsection{Main Results}
\begin{table*}[t]
	\begin{center}
	\resizebox{2.08\columnwidth}{!}{
		\begin{tabular}{c|ccccccccccccc|cc}
			\hline
			\hline
			&\multirow{2}{*}{Methods}&SST-2&Yelp P.&AG's News&DBPedia&MRPC&SNLI&RTE&\multirow{2}{*}{Average} \\
            &&acc&acc&acc&acc&F1&acc&acc \\
            \hline
            \specialrule{0em}{1pt}{1pt}
			\multirow{6}{*}{Gradient-based}
			&Model Tuning
			& 85.39\scriptsize $\pm$2.84 
            & 91.82\scriptsize $\pm$0.79 
			& 86.36\scriptsize $\pm$1.85      
			& 97.98\scriptsize $\pm$0.14 
			& 77.35\scriptsize $\pm$5.70
			& 54.64\scriptsize $\pm$5.29
			& 58.60\scriptsize $\pm$6.21
			& 78.88 \\
			\specialrule{0em}{1pt}{1pt}
			&Adapter
			& 83.91\scriptsize $\pm$2.90 
            & 90.99\scriptsize $\pm$2.86 
			& 86.01\scriptsize $\pm$2.18      
			& \textbf{97.99\scriptsize $\pm$0.07}
			& 69.20\scriptsize $\pm$3.58
			& 57.46\scriptsize $\pm$6.63
			& 48.62\scriptsize $\pm$4.74
			& 76.31 \\
			\specialrule{0em}{1pt}{1pt}
			&BitFit
			& 81.19\scriptsize $\pm$6.08 
            & 88.63\scriptsize $\pm$6.69 
			& 86.83\scriptsize $\pm$0.62      
			& 94.42\scriptsize $\pm$0.94 
			& 66.26\scriptsize $\pm$6.81
			& 53.42\scriptsize $\pm$10.63
			& 52.59\scriptsize $\pm$5.31
			& 74.76 \\
			\specialrule{0em}{1pt}{1pt}
			&LoRA
			& 88.49\scriptsize $\pm$2.90 
            & 90.21\scriptsize $\pm$4.00 
			& 87.09\scriptsize $\pm$0.85      
			& 97.86\scriptsize $\pm$0.17 
			& 72.14\scriptsize $\pm$2.23
			& 61.03\scriptsize $\pm$8.55
			& 49.22\scriptsize $\pm$5.12
			& 78.01 \\
			\specialrule{0em}{1pt}{1pt}
			&Prompt Tuning
			& 68.23\scriptsize $\pm$3.78 
            & 61.02\scriptsize $\pm$6.65 
			& 84.81\scriptsize $\pm$0.66      
			& 87.75\scriptsize $\pm$1.48 
			& 51.61\scriptsize $\pm$8.67
			& 36.13\scriptsize $\pm$1.51
			& 54.69\scriptsize $\pm$3.79
			& 63.46 \\
			\specialrule{0em}{1pt}{1pt}
			&P-Tuning v2
			& 64.33\scriptsize $\pm$3.05 
            & 92.63\scriptsize $\pm$1.39 
			& 83.46\scriptsize $\pm$1.01      
			& 97.05\scriptsize $\pm$0.41 
			& 68.14\scriptsize $\pm$3.89
			& 36.89\scriptsize $\pm$0.79
			& 50.78\scriptsize $\pm$2.28
			& 70.47 \\
			\specialrule{0em}{1pt}{1pt}
			\hline
			\specialrule{0em}{1pt}{1pt}
			\multirow{6}{*}{Gradient-free}
			&Manual Prompt
			& 79.82 & 89.65 & 76.96 & 41.33 & 67.40 & 31.11 & 51.62 & 62.56 \\
			\specialrule{0em}{1pt}{1pt}
			&In-Context Learning
			& 79.79\scriptsize $\pm$3.06 
            & 85.38\scriptsize $\pm$3.92 
			& 62.21\scriptsize $\pm$13.46      
			& 34.83\scriptsize $\pm$7.59 
			& 45.81\scriptsize $\pm$6.67
			& 47.11\scriptsize $\pm$0.63
			& 60.36\scriptsize $\pm$1.56
			& 59.36 \\
			\specialrule{0em}{1pt}{1pt}
			&Feature-MLP
			& 64.80\scriptsize $\pm$1.78 
            & 79.20\scriptsize $\pm$2.26 
			& 70.77\scriptsize $\pm$0.67      
			& 87.78\scriptsize $\pm$0.61 
			& 68.40\scriptsize $\pm$0.86
			& 42.01\scriptsize $\pm$0.33
			& 53.43\scriptsize $\pm$1.57
			& 66.63 \\
			\specialrule{0em}{1pt}{1pt}
			&Feature-BiLSTM
			& 65.95\scriptsize $\pm$0.99 
            & 74.68\scriptsize $\pm$0.10 
			& 77.28\scriptsize $\pm$2.83      
			& 90.37\scriptsize $\pm$3.10 
			& 71.55\scriptsize $\pm$7.10
			& 46.02\scriptsize $\pm$0.38
			& 52.17\scriptsize $\pm$0.25
			& 68.29 \\
			\specialrule{0em}{1pt}{1pt}
			&BBT
			& 89.56\scriptsize $\pm$0.25 
            & 91.50\scriptsize $\pm$0.16 
			& 81.51\scriptsize $\pm$0.79      
			& 79.99\scriptsize $\pm$2.95 
			& 61.56\scriptsize $\pm$4.34
			& 46.58\scriptsize $\pm$1.33
			& 52.59\scriptsize $\pm$2.21
			& 71.90 
            \\
            \specialrule{0em}{1pt}{1pt}
			&BBTv2
			& 90.33\scriptsize $\pm$1.73 
            & 92.86\scriptsize $\pm$0.62 
			& 85.28\scriptsize $\pm$0.49      
			& 93.64\scriptsize $\pm$0.68 
			& 77.01\scriptsize $\pm$4.73
			& 57.27\scriptsize $\pm$2.27
			& 56.68\scriptsize $\pm$3.32
			& 79.01 \\
			\hline
			\specialrule{0em}{1pt}{1pt}
			Hybrid
			&\textbf{GDFO~(ours)}
			& \textbf{92.41\scriptsize $\pm$1.03}
            & \textbf{93.17\scriptsize $\pm$0.37}
			& \textbf{87.19\scriptsize $\pm$0.51}   
			& 96.92\scriptsize $\pm$0.71
			& \textbf{80.13\scriptsize $\pm$1.97}
			& \textbf{62.53\scriptsize $\pm$1.31}
			& \textbf{60.57\scriptsize $\pm$1.02}
			& \textbf{81.85}  \\
			\hline
			\hline
		\end{tabular}
		}
	\end{center}
	\caption{Results~(\%) of 16-shot setting on various downstream tasks. Following \citet{sun2022bbtv2}, we report mean and standard deviation of performance over 3 different splits.We highlight the best results in \textbf{bold}.}
	\label{tab main experiment}
	
\end{table*}

The results
of 16-shot setting 
on various downstream tasks
are shown in Table~\ref{tab main experiment}.
From the table,
GDFO consistently outperforms 
all the baselines on the average performance.
Specifically,
in the gradient-based comparison,
GDFO achieves an
average accuracy of $81.85\%$,
which outperforms the runner-up 
gradient-based model,
LoRA,
by a notable $3.84\%$ improvement.
When compared against the gradient-free methods,
GDFO leads BBTv2 by $5.26\%$ and $3.89\%$ 
on the SNLI and RTE datasets,
respectively.
Our model generates a continuous prompt 
for each sample, 
rather than using an optimized continuous prompt
for all samples,
such as BBT and BBTv2.
Furthermore,
the incorporation of 
both DFO and gradient descent
during the training stage
allows GDFO for more comprehensive 
and efficient training of continuous prompts,
resulting in a notable improvement 
in the model performance.


\subsection{Ablation Study}
\label{sec:ablation}

\begin{figure*}[htbp]
    \centering
    \includegraphics[height=105pt, width=455pt]{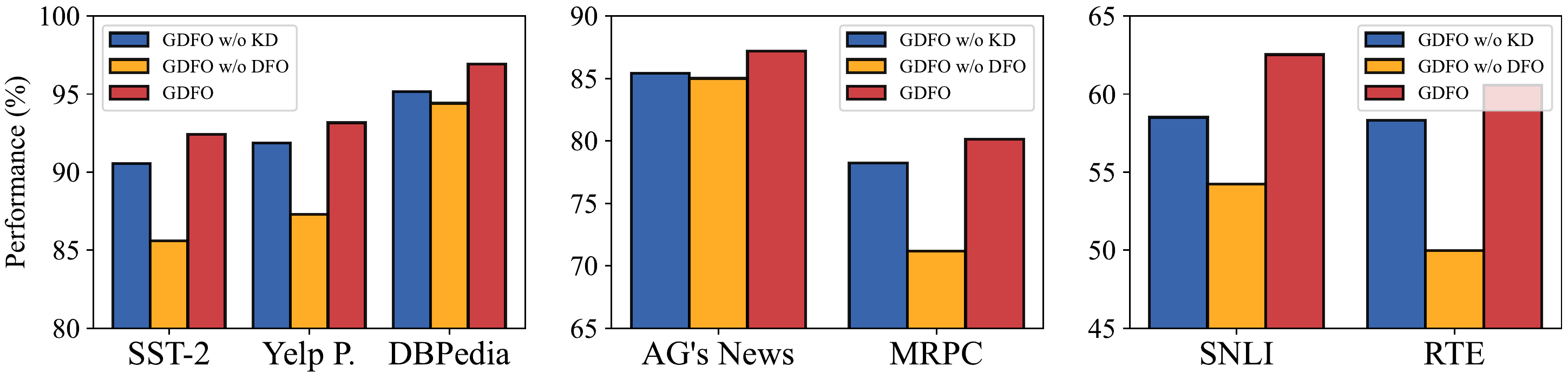}
    \caption{Ablation study: 
    Results~(\%) of 16-shot problems over seven datasets.
    \textbf{w/o KD} denotes removing knowledge distillation 
    and \textbf{w/o DFO} denotes removing derivative-free optimization.
    When removing the prompt generator, our method degrades to BBT~\cite{sun2022bbt}. The comparison of GDFO and BBT is shown in Table~\ref{tab main experiment}. The detailed analysis are described in Section~\ref{sec:ablation}.}
    \label{fig:ablation}
\end{figure*}

We conduct an ablation study 
to investigate the characteristics 
of the main components of GDFO.
As illustrated in Figure~\ref{fig:ablation},
the results\footnote{The evaluation metric 
used in the ablation study
is F1 score for MRPC 
and accuracy for other datasets.} 
demonstrate that
GDFO outperforms GDFO-w/o-KD. 
For instance,
on the SNLI dataset,
the accuracy of GDFO is $62.53\%$,
whereas that of GDFO-w/o-KD is only $58.51\%$.
This indicates that 
the knowledge distillation module,
which transfers the knowledge
of the \emph{teacher model} to the \emph{student model}
by aligning the outputs of the \emph{student model} 
with those of the \emph{teacher model}, 
effectively improves the model performance.
Additionally,
when removing derivative-free optimization,
a significant decline is observed across all datasets,
with an average decrease of $6.5\%$.
This demonstrates the effectiveness of
incorporating derivative-free optimization in the black-box scenario.
It is worth noting that 
when removing the prompt generator,
the student model will not function, 
which means that gradient descent is eliminated.
In this case,
our method degrades to a gradient-free method~BBT.
The results, as shown in Table~\ref{tab main experiment},
reveal that
GDFO
achieves
significant performance gains 
over BBT
across
all datasets,
which
demonstrates
the effectiveness of 
training the prompt generator 
through gradient descent in the black-box scenarios.


\subsection{Analysis}

\paragraph{Different Black-Box Models}

\begin{figure}[!t]
    \centering
    \includegraphics[width=0.48\textwidth]{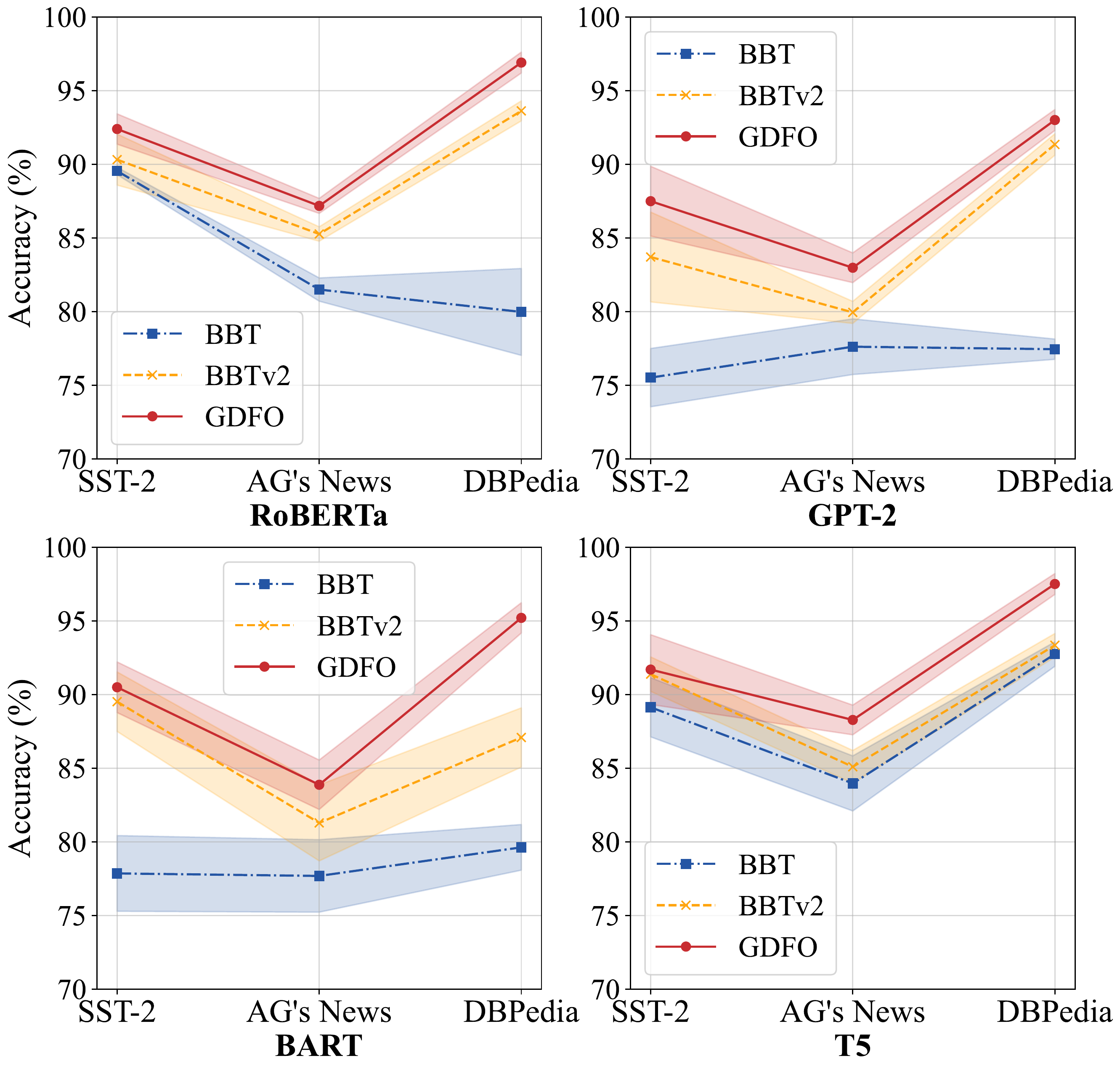}
    \caption{Accuracy~(\%) on different black-box models.
    We report mean and standard deviation of performance 
    over 3 different splits.
    The results of BBT and BBTv2 are reported in~\cite{sun2022bbtv2}.}
    \label{fig:different-bb}
\end{figure}

To evaluate the universality of GDFO across PLMs
with varying architectures,
in addition to 
encoder-only PLMs~(e.g., $\text{RoBERTa}_\text{{\footnotesize LARGE}}$),
we conduct experiments using
decoder-only~(e.g., $\text{GPT-2}_\text{{\footnotesize LARGE}}$)
and encoder-decoder PLMs~(e.g., $\text{BART}_\text{{\footnotesize LARGE}}$
and 
$\text{T5}_\text{{\footnotesize LARGE}}$) 
as black-box models.
As shown in Figure~\ref{fig:different-bb},
GDFO achieves superior performance over
other competitors across all the settings.
For example,
When using GPT-2 as the black-box model,
GDFO achieves $87.5\%$ and $85.2\%$
on the SST-2 and DBPedia datasets,
respectively.
In particular,
it outperforms BBT
by a notable $11.9\%$ and $15.5\%$
improvements in both cases.
When considering BART as the black-box model,
GDFO leads BBTv2 by $8.12\%$
on the DBPedia dataset.
All the results clearly show 
the generalizability of our model across various PLMs.

\paragraph{Different Student Models}

We next conduct an in-depth experiment 
for student models on three datasets.
The results are shown in Table~\ref{tab:StuModel}.
From the results,
different student models
have a impact on the performance of GDFO
(approximately $2\%$ on average).
Furthermore, 
we observe that
student models 
whose architectures are similar 
to the black-box model 
tend to 
exhibit superior performance.
For instance,
when both the black-box model 
and the student model are
$\text{RoBERTa}_\text{{\footnotesize LARGE}}$~\cite{liu2019roberta},
GDFO achieves
the best performance.
When comparing models 
with identical architectures, 
such as $\text{BART}_\text{{\footnotesize LARGE}}$~\cite{lewis2020bart}
and 
$\text{T5}_\text{{\footnotesize LARGE}}$~\cite{raffel2020_t5}, T5 exhibits superior performance,
which may be due to the fact that 
the T5 model has twice the number of parameters
as the BART model. 
The increased capacity 
allows the T5 model 
to better capture and 
represent the relationships 
within the input data, 
resulting in improved performance.

\begin{table}[!t]
	\begin{center}
	\resizebox{1\columnwidth}{!}{
		\begin{tabular}{cccc}
			\hline
			\hline
			\specialrule{0em}{1pt}{1pt}
			Stu. Models & SST-2 & AG's News & DBPedia \\
            \hline
            \specialrule{0em}{1pt}{1pt}
            \multicolumn{4}{c}{\textcolor{gray}{\textit{Encoder-only PLMs}}} \\
            \hline
            \specialrule{0em}{1pt}{1pt}
            $\text{BERT}_\text{{\scriptsize LARGE}}$
            & 92.41\scriptsize $\pm$1.03
			& 87.19\scriptsize $\pm$0.51
			& 96.92\scriptsize $\pm$0.71\\
			\specialrule{0em}{1pt}{1pt}
            $\text{RoBERTa}_\text{{\scriptsize LARGE}}$
            & \textbf{93.17\scriptsize $\pm$0.39}
			& \textbf{88.91\scriptsize $\pm$0.47}
			& \textbf{97.56\scriptsize $\pm$0.53}\\
            \hline
			\specialrule{0em}{1pt}{1pt}
            \multicolumn{4}{c}{\textcolor{gray}{\textit{Decoder-only PLMs}}} \\
            \hline
            \specialrule{0em}{1pt}{1pt}
            $\text{GPT-2}_\text{{\scriptsize LARGE}}$
            & 91.12\scriptsize $\pm$1.72
			& 85.98\scriptsize $\pm$1.28
			& 95.91\scriptsize $\pm$2.01\\
            \hline
            \specialrule{0em}{1pt}{1pt}
            \multicolumn{4}{c}{\textcolor{gray}{\textit{Encoder-Decoder PLMs}}} \\
            \hline
            \specialrule{0em}{1pt}{1pt}
            $\text{BART}_\text{{\scriptsize LARGE}}$
            & 91.19\scriptsize $\pm$0.93
			& 87.07\scriptsize $\pm$0.57
			& 96.13\scriptsize $\pm$0.82\\
			\specialrule{0em}{1pt}{1pt}
			$\text{T5}_\text{{\scriptsize LARGE}}$
            & 93.03\scriptsize $\pm$0.31
			& 88.87\scriptsize $\pm$0.47
			& 97.73\scriptsize $\pm$0.98\\
			\hline
			\hline
		\end{tabular}
		}
	\end{center}
	\caption{Accuracy~(\%) of GDFO based on different student models. The black-box model is $\text{RoBERTa}_\text{{\scriptsize LARGE}}$~\cite{liu2019roberta}.
	We report mean and standard deviation of performance 
	over 3 different splits.
	We highlight the best results in bold.}
	\label{tab:StuModel}
	
\end{table}

\paragraph{Effect of Balancing Weight}

\begin{figure}[!t]
    \centering
    \includegraphics[width=0.47\textwidth]{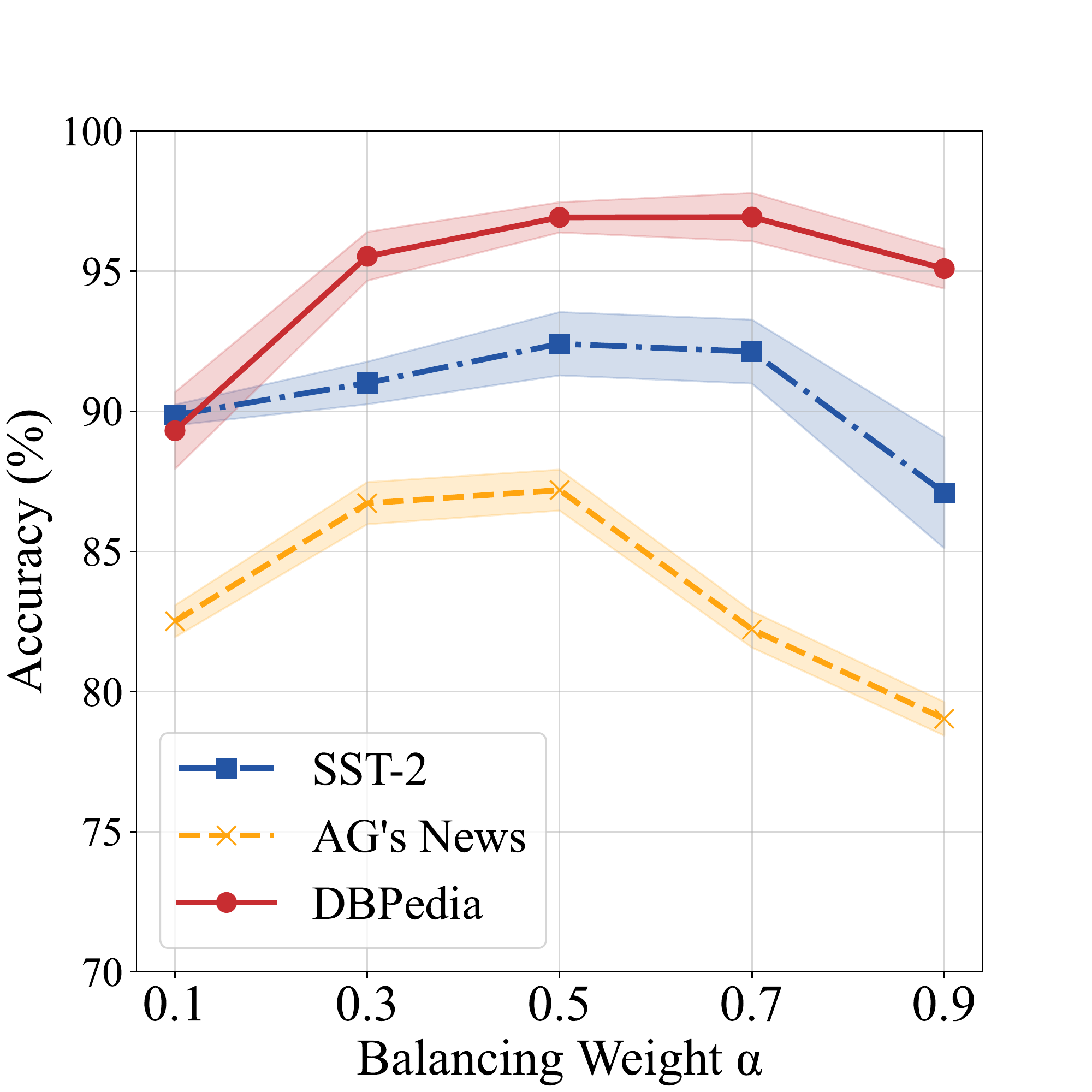}
    \caption{Effect of the balancing weight~$\alpha$ on three datasets.
    We report mean and standard deviation of performance 
    over 3 different splits.}
    \label{fig:hyper}
\end{figure}

The balancing weight $\alpha$ 
plays a crucial role 
in determining the performance of the model 
by controlling the influence of 
$\mathbf{p}_{\text{\tiny GD}}$ 
and 
$\mathbf{Az}$. 
As the value of $\alpha$ increases, 
the influence of $\mathbf{p}{\text{\tiny GD}}$ 
becomes more prominent,
while conversely, 
as the value of $\alpha$ decreases,
the influence of $\mathbf{Az}$ 
becomes more pronounced\footnote{$\mathbf{p}_0$ is fixed,
thus its effect on
the model performance
is disregarded in the analysis.}.
As illustrated in the Figure~\ref{fig:hyper}, 
when $\alpha$ is set to an extreme value, 
either too large or too small, 
it tends to
have a negative impact
on the model performance.
We observe that
the average performance of the model across three datasets
is optimal when $\alpha$ is set to $0.5$,
further emphasizing the importance of 
the combination of derivative-free optimization 
and gradient descent in improving the performance of the model.

\section{Conclusion}
In this paper,
we introduced gradient descent into the black-box tuning scenario 
through knowledge distillation for the first time,
which provided a novel insight
for future black-box tuning approaches.
Furthermore,
we proposed a novel method,
GDFO,
which integrates
gradient descent 
and derivative-free optimization
for jointly training continuous prompts.
GDFO first trains a \emph{student model} 
to enhance the performance 
by aligning its outputs 
with those of 
the \emph{teacher model}~(i.e., the black-box model).
After that,
GDFO 
trains a prompt generator using gradient descent
while simultaneously optimizing
a continuous prompt using DFO algorithm.
Experimental results on various datasets
show that 
GDFO can achieve significant performance gains
over other gradient-free and gradient-based methods.

\section*{Limitations}
We summarize the limitations of this work as follows:
(1) We conduct experiments 
on 7 language understanding tasks
across 4 types~(i.e., sentiment analysis,
topic classification, 
natural language inference
and paraphrasing). 
However, the effectiveness of GDFO
on tasks such as sequence labeling
and generation tasks
has yet to be fully examined.
(2) Our proposed method 
uses a student model
and a prompt generator,
thereby resulting 
in a higher computational resource requirement
in comparison to gradient-free methods.
Therefore, 
it may not be suitable for 
implementation on certain edge devices,
but it is more appropriate 
for personal or enterprise users
who have access to a certain degree
of computational resources
and have stringent requirements
for the model performance.
(3) We only focus on the few-shot setting
in this paper.
It is possible to 
extend our work to other scenarios 
such as semi-supervised learning 
and we will further explore it 
in the future research.

\section*{Ethics Statement}
The proposed method has no obvious potential risks. 
All the scientific artifacts used/created 
are properly cited/licensed, 
and the usage is consistent with 
their intended use. 

\section*{Acknowledgements}
\label{sec:acknowledge}
This work has been supported by the National Natural Science Foundation of China under Grant No. U1911203, 
and the National Natural Science Foundation of China under Grant No.61977025.

\bibliography{custom}
\bibliographystyle{acl_natbib}




\end{document}